\documentclass[runningheads]{llncs}

\usepackage{graphicx}
\usepackage{color}

\usepackage{multirow}
\usepackage{xspace}
\makeatletter
\DeclareRobustCommand\onedot{\futurelet\@let@token\@onedot}
\def\@onedot{\ifx\@let@token.\else.\null\fi\xspace}

\def\ie{\emph{i.e}\onedot}

\def\etal{\emph{et al}\onedot}
\makeatother

\newcommand\Tstrut{\rule{0pt}{2.6ex}} 

\begin{document}

\title{Deep Object Co-Segmentation} 
\titlerunning{Deep Object Co-Segmentation} 


\newcommand*\samethanks[1][\value{footnote}]{\footnotemark[#1]}

\author{Weihao Li\thanks{Equal contribution} \and
Omid Hosseini Jafari\samethanks \and 
Carsten Rother}
%
\index{Hosseini Jafari, Omid}

\authorrunning{W. Li, O. Hosseini Jafari, C. Rother} 

\institute{Visual Learning Lab, Heidelberg University (HCI/IWR) \\
(weihao.li, omid.hosseini\_jafari, carsten.rother)@iwr.uni-heidelberg.de
}

\maketitle

\begin{abstract}
This work presents a deep object co-segmentation (DOCS) approach for 
segmenting common objects of the same class within a pair of images. 
This means that the method learns to ignore common, or uncommon, 
background \textit{stuff} and focuses on common \textit{objects}. 
If multiple object classes are presented in the image pair, 
they are jointly extracted as foreground. To address this task, 
we propose a CNN-based Siamese encoder-decoder architecture. 
The encoder extracts high-level semantic features of the foreground objects, 
a mutual correlation layer detects the common objects, 
and finally, the decoder generates the output foreground masks for each image. 
To train our model, we compile a large object co-segmentation dataset consisting of 
image pairs from the PASCAL dataset with common objects masks. 
We evaluate our approach on commonly used datasets for co-segmentation tasks 
and observe that our approach consistently outperforms competing methods, 
for both seen and unseen object classes.
\end{abstract}

\begin{figure*}[t]
\begin{center}
\includegraphics[width=0.90\linewidth]{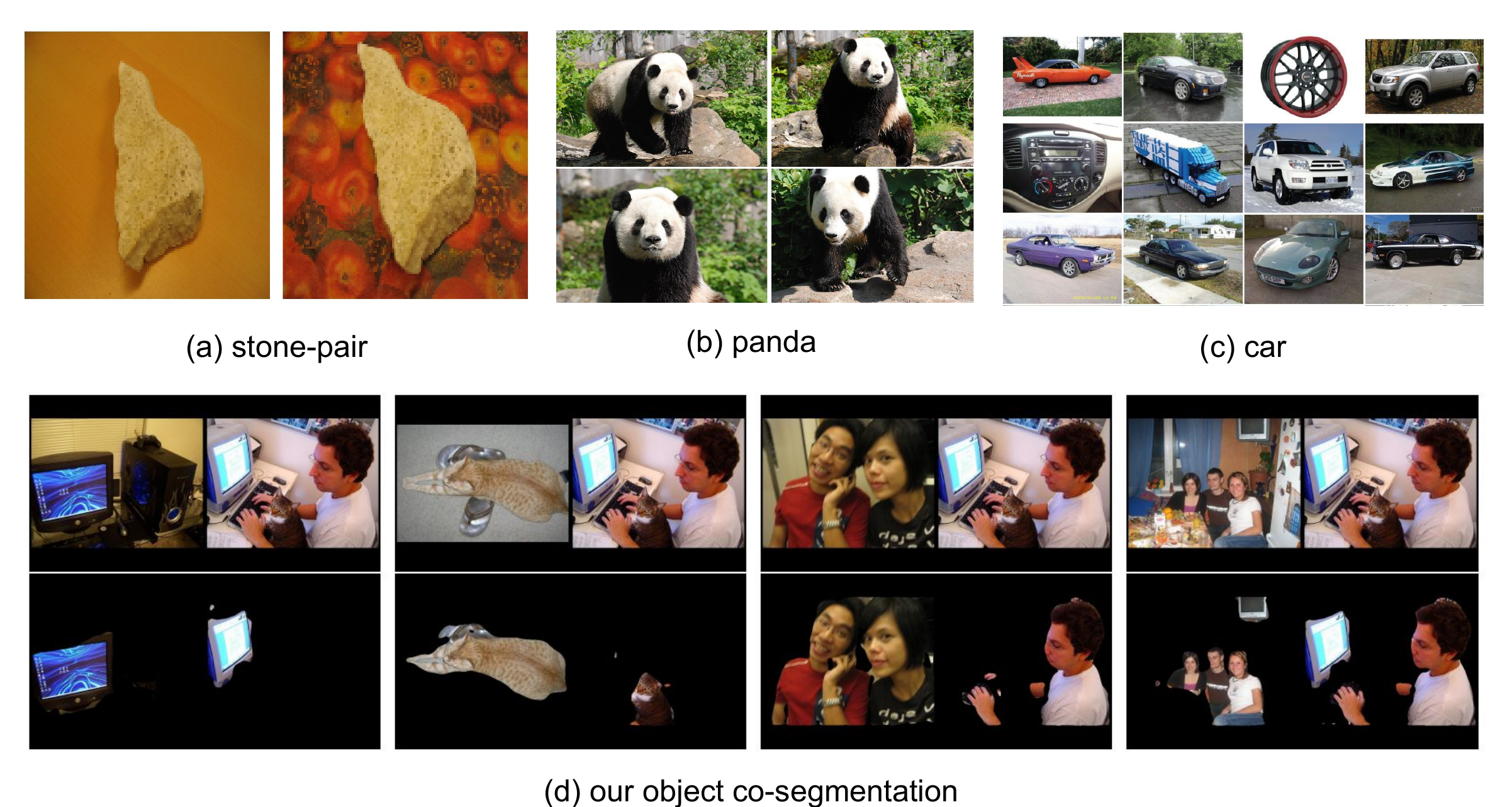}
\end{center}
   \caption{\textbf{Different co-segmentation challenges:} 
   (a) segmenting common parts, in terms of small appearance deviation,
   with varying background \cite{rother2006cosegmentation}, 
   (b) segmenting common objects from the same class with low intra-class variation 
   but similar background \cite{batra2010icoseg,vicente2010cosegmentation}, 
   (c) segmenting common objects from the same class with large variability in terms of 
   scale, appearance, pose, viewpoint and background \cite{Rubinstein_2013_ICCV}. 
   (d) segmenting common objects in images including more than one object from multiple classes. Second row shows our predicted co-segmentation of these challenging images.
}
\label{fig:example}
\end{figure*}

\section{Introduction}
\label{introduction}
Object co-segmentation is the task of segmenting 
the common objects from a set of images. It is applied 
in various computer vision applications and beyond, 
such as browsing in photo collections \cite{rother2006cosegmentation}, 
3D reconstruction \cite{kowdle2010imodel}, 
semantic segmentation \cite{shen2017weakly}, 
object-based image retrieval \cite{vicente2011object}, 
video object tracking and segmentation \cite{rother2006cosegmentation}, 
and interactive image segmentation \cite{rother2006cosegmentation}. 

There are different challenges for object co-segmentation 
with varying level of difficulty: 
(1) Rother \etal \cite{rother2006cosegmentation} 
first proposed the term of \textit{co-segmentation} as the task of 
segmenting the \textit{common parts} of an image pair simultaneously. 
They showed that segmenting two images jointly achieves 
better accuracy in contrast to segmenting them independently. 
They assume that the common parts have similar appearance. 
However, the background in both images are significantly different, 
see Fig.~\ref{fig:example}(a).
(2) Another challenge is to segment the same object instance 
or similar objects of the same class with low intra-class variation, 
even with similar background \cite{batra2010icoseg,vicente2011object}, 
see Fig.~\ref{fig:example}(b). 
(3) A more challenging task is to segment common objects from the same class 
with large variability in terms of scale, appearance, pose, viewpoint 
and background \cite{Rubinstein_2013_ICCV}, see Fig.~\ref{fig:example}(c). 

All of the mentioned challenges assume that the image set contains only one common object 
and the common object should be salient within each image. In this work, we address a more
general problem of co-segmentation without this assumption, \ie multiple object classes can be 
presented within the images, see Fig.~\ref{fig:example}(d).
As it is shown, the co-segmentation result for one specific image including multiple objects 
can be different when we pair it with different images.
Additionally, we are interested in co-segmenting objects, \ie \textit{things} rather than \textit{stuff}. 
The idea of object co-segmentation was introduced by Vicente \etal \cite{vicente2011object}
to emphasize the resulting segmentation to be a \textit{thing} such as a `cat' or a `monitor', 
which excludes common, or uncommon, \textit{stuff} classes like `sky' or `sea'. 

Segmenting objects in an image is one of the fundamental tasks in 
computer vision. While image segmentation has received great attention 
during the recent rise of deep learning 
\cite{Long_2015_CVPR,U-net,Zheng_2015_ICCV,Xu_2016_CVPR,Quan_2016_CVPR}, 
the related task of object co-segmentation remains largely unexplored 
by newly developed deep learning techniques. Most of the recently 
proposed object co-segmentation methods still rely on models without 
feature learning. This includes methods utilizing super-pixels, 
or proposal segments \cite{vicente2011object,Taniai_2016_CVPR} to 
extract a set of object candidates, or methods which use a complex 
CRF model \cite{Lee_2015_CVPR,Quan_2016_CVPR} with hand-crafted features 
\cite{Quan_2016_CVPR} to find the segments with the highest similarity. 

In this paper, we propose a simple yet powerful method for segmenting 
objects of a common semantic class from a pair of images using a  
convolutional encoder-decoder neural network. Our method uses a pair 
of Siamese encoder networks to extract semantic features for each image. 
The mutual correlation layer at the network's bottleneck computes localized 
correlations between the semantic features of the two images to highlight 
the heat-maps of common objects. Finally, the Siamese decoder networks combine 
the semantic features from each image with the correlation features to 
produce detailed segmentation masks through a series of deconvolutional 
layers. Our approach is trainable in an end-to-end manner and does not 
require any, potentially long runtime, CRF optimization procedure at evaluation time. 
We perform an extensive evaluation of our deep object co-segmentation 
and show that our model can achieve state-of-the-art performance on 
multiple common co-segmentation datasets. In summary, our main contributions are as follows:
\begin{itemize}
\item We propose a simple yet effective convolutional neural network (CNN)
	  architecture for object co-segmentation that can be trained end-to-end.
      To the best of our knowledge, this is the first pure
      CNN framework for object co-segmentation, 
      which does not depend on any hand-crafted features.
\item We achieve state-of-the-art results on multiple object co-segmentation datasets, 
      and introduce a challenging object co-segmentation dataset by adapting Pascal
      dataset for training and testing object co-segmentation models.
\end{itemize}

\section{Related Work}
\label{sec:relatedwork}

We start by discussing object co-segmentation by roughly categorizing 
them into three branches: co-segmentation without explicit learning,  
co-segmentation with learning, and interactive co-segmentation. 
After that, we briefly discuss various image segmentation tasks 
and corresponding approaches based on CNNs.

\noindent \textbf{Co-Segmentation without Explicit Learning.}
Rother \etal \cite{rother2006cosegmentation} proposed the problem of 
image co-segmentation for image pairs. They minimize an energy function 
that combines an MRF smoothness prior term with a histogram matching term. 
This forces the histogram statistic of common foreground regions 
to be similar. In a follow-up work,  
Mukherjee \etal \cite{mukherjee2009half} replace the $l_1$ norm 
in the cost function by an $ l_2 $ norm.
In \cite{hochbaum2009efficient}, Hochbaum and Singh used 
a reward model, in contrast to the penalty strategy of 
\cite{rother2006cosegmentation}.
In \cite{vicente2010cosegmentation}, Vicente \etal studied 
various models and showed that a simple model based on Boykov-Jolly \cite{boykov2001interactive} works the best.
Joulin \etal \cite{joulin2010discriminative} formulated the 
co-segmentation problem in terms of a discriminative clustering task.
Rubio \etal \cite{rubio2012unsupervised} proposed to match regions, 
which results from an over-segmentation algorithm, 
to establish correspondences between the common objects. 
Rubinstein \etal \cite{Rubinstein_2013_ICCV} combined a visual saliency 
and dense correspondences, using SIFT flow, to capture the sparsity 
and visual variability of the common object in a group of images. 
Fu \etal \cite{Fu_2015_CVPR} formulated object co-segmentation 
for RGB-D input images as a fully-connected graph structure, 
together with mutex constraints.
In contrast to these works, our method is a pure learning based approach.

\noindent \textbf{Co-Segmentation with Learning.}
In \cite{vicente2011object}, Vicente \etal generated a pool of 
object-like proposal-segmentations using constrained parametric min-cut 
\cite{carreira2010constrained}. Then they trained a random forest 
classifier to score the similarity of a pair of segmentation proposals. 
Yuan \etal \cite{DDCRF} introduced a deep dense conditional random field 
framework for object co-segmentation by inferring co-occurrence maps. 
These co-occurrence maps measure the objectness scores, as well as, 
similarity evidence for object proposals, which are generated using 
selective search \cite{uijlings2013selective}. 
Similar to the constrained parametric min-cut, selective search also 
uses hand-crafted SIFT and HOG features to generate object proposals. 
Therefore, the model of \cite{DDCRF} cannot be trained end-to-end. 
In addition, \cite{DDCRF} assume that there is a single common object 
in a given image set, which limits application in real-world scenarios. 
Recently, Quan \etal \cite{Quan_2016_CVPR} proposed a manifold ranking 
algorithm for object co-segmentation by combining low-level appearance 
features and high-level semantic features. However, their semantic features
are pre-trained on the ImageNet dataset. 
In contrast, our method is based on a pure CNN architecture, 
which is free of any hand-crafted features and object proposals
and does not depend on any assumption about the existence of common objects.

\noindent \textbf{Interactive Co-Segmentation.} 
Batra \etal \cite{batra2010icoseg} firstly presented an algorithm 
for interactive co-segmentation of a foreground object from a group 
of related images. They use users' scribbles to indicate the foreground.
Collins \etal \cite{collins2012random} used a random walker model 
to add consistency constraints between  foreground regions within 
the interactive co-segmentation framework. However, their co-segmentation 
results are sensitive to the size and positions of users' scribbles.
Dong \etal \cite{dong2015interactive} proposed an interactive 
co-segmentation method which uses global and local energy optimization, 
whereby the energy function is based on scribbles, inter-image 
consistency, and a standard local smoothness prior.
In contrast, our work is not a user-interactive co-segmentation approach.

\noindent \textbf{Convolutional Neural Networks for Image Segmentation.}
In the last few years, CNNs have achieved 
great success for the tasks of image segmentation, 
such as semantic segmentation 
\cite{Long_2015_CVPR,Noh_2015_ICCV,YuKoltun2016,Lin_2017_CVPR,Xu_2016_CVPR,Zhao_2017_CVPR},
interactive segmentation \cite{Xu_2016_CVPR,xu2017deep}, 
and salient object segmentation \cite{Li_2016_CVPR,wang2016saliency,jain2017pixel}. 

Semantic segmentation aims at assigning semantic labels to each pixel in an image. 
Fully convolutional networks (FCN) \cite{Long_2015_CVPR} became one of the 
first popular architectures for semantic segmentation. 
Nor \etal \cite{Noh_2015_ICCV} proposed a deep deconvolutional network 
to learn the upsampling of low-resolution features.
Both U-Net \cite{U-net} and SegNet \cite{badrinarayanan2017segnet} 
proposed an encoder-decoder architecture, in which the decoder network 
consists of a hierarchy of decoders, each corresponding to an encoder.
Yu \emph{et al.} \cite{YuKoltun2016} and Chen \emph{et al.} \cite{chen14semantic} 
proposed dilated convolutions to aggregate multi-scale contextual information, 
by considering larger receptive fields.
Salient object segmentation aims at detecting and segmenting the 
salient objects in a given image. Recently, deep learning architectures 
have become popular for salient object segmentation 
\cite{Li_2016_CVPR,wang2016saliency,jain2017pixel}. 
Li and Yu \cite{Li_2016_CVPR} addressed salient object segmentation 
using a deep network which consists of a pixel-level multi-scale 
FCN and a segment scale spatial pooling stream. 
Wang \etal \cite{wang2016saliency} proposed recurrent FCN to incorporate saliency prior knowledge for improved inference, utilizing a pre-training strategy based on semantic segmentation data.  
Jain \etal \cite{jain2017pixel} proposed to train a  
FCN to produce pixel-level masks 
of all “object-like” regions given a single input image. 

Although CNNs play a central role in 
image segmentation tasks, there has been no prior work with 
a pure CNN architecture for object co-segmentation. 
To the best of our knowledge, our deep CNN
architecture is the first of its kind for object co-segmentation.  

\begin{figure*}[t]
\begin{center}
\includegraphics[width=\linewidth]{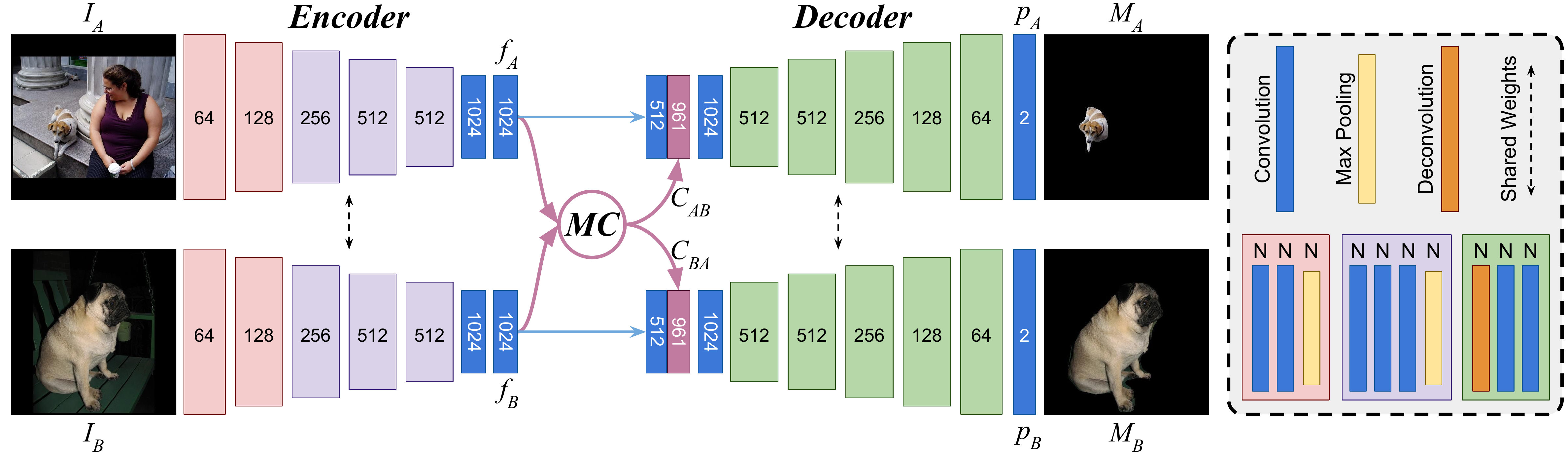}
\end{center}
   \caption{\textbf{Deep Object Co-Segmentation}. Our network includes three parts: 
   (i) passing input images $I_A$ and $I_B$ through a Siamese encoder  
   to extract feature maps $f_A$ and $f_B$, 
   (ii) using a mutual correlation network to perform feature matching to obtain 
   correspondence maps $C_{AB}$ and $C_{BA}$, 
   (iii) passing concatenation of squeezed feature maps and correspondence maps
   through a Siamese decoder to get the common objects masks $M_A$ and $M_B$.}
\label{fig:cnn_docs}
\end{figure*}

\section{Method}
\label{sec:method}

In this section, we introduce a new CNN 
architecture for segmenting the common objects from two input images. 
The architecture is end-to-end trainable for the object co-segmentation 
task. Fig. \ref{fig:cnn_docs} illustrates the overall structure of our 
architecture. Our network consists of three main parts: (1) Given two input 
images $I_A$ and $I_B$, we use a Siamese encoder to extract high-level semantic feature maps $f_A$ and $f_B$. 
(2) Then, we propose a mutual correlation layer to obtain correspondence 
maps $C_{AB}$ and $C_{BA}$ by matching feature maps $f_A$ and $f_B$ at 
pixel-level. (3) Finally, given the concatenation of the feature maps 
$f_A$ and $f_B$ and correspondence maps $C_{AB}$ and $C_{BA}$, a Siamese 
decoder is used to obtain and refine the common object masks $M_A$ and $M_B$. 

In the following, we first describe each of the three parts of 
our architecture in detail. Then in Sec~\ref{sub_sec::loss}, the 
loss function is introduced. Finally, in Sec~\ref{sub_sec::coseg_group}, 
we explain how to extend our approach to handle co-segmentation 
of a group of images, \ie going beyond two images.

\subsection{Siamese Encoder}
The first part of our architecture is a Siamese encoder which consists 
of two identical feature extraction CNNs with shared parameters. 
We pass the input image pair $I_A$ and $I_B$ through the Siamese 
encoder network pair to 
extract feature maps $f_A$ and $f_B$. More specifically, our encoder 
is based on the VGG16 network~\cite{VGG}. We keep the first 
13 convolutional layers and 
replace \textit{fc6} and \textit{fc7} with two  
$3 \times 3 $ convolutional layers \textit{conv6-1} and \textit{conv6-2} 
to produce feature maps which contain more spatial information. In total, 
our encoder network has 15 convolutional layers and 5 pooling layers 
to create a set of high-level semantic features $f_A  $ and $f_B$.
The input to the Siamese encoder is two $512 \times 512$ images and 
the output of the encoder is two $1024$-channel feature maps with a 
spatial size of $16 \times 16$.

\subsection{Mutual Correlation}
\begin{figure}[t]
	\centering
	\includegraphics[width=0.90\linewidth]{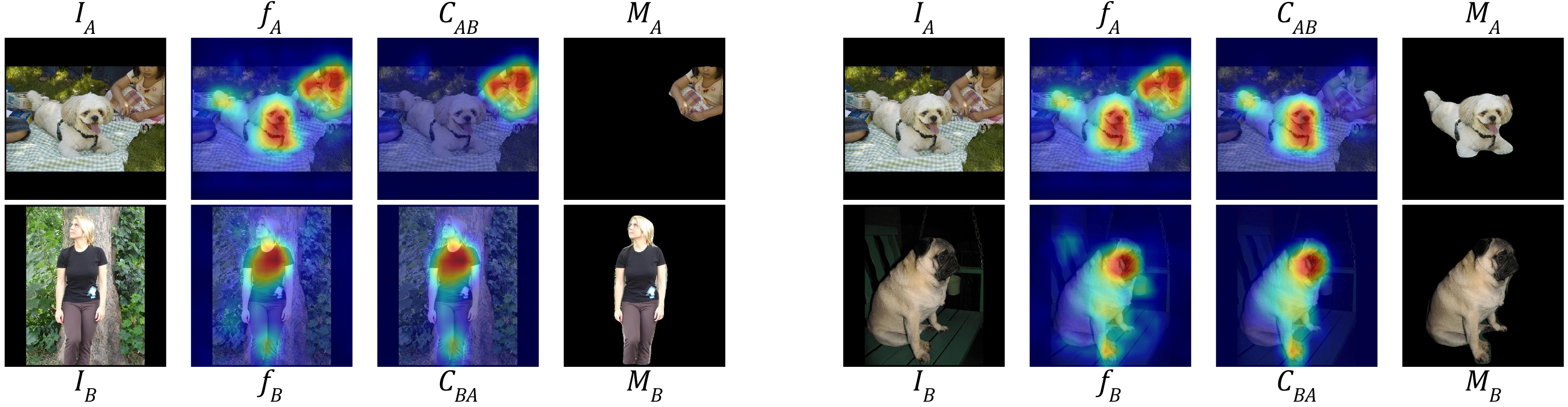}
	\caption{\textbf{The visualization of the heat-maps.} 
    Given a pair of input images $I_A$ and $I_B$, after passing them 
    through the Siamese encoder, 
    we extract feature maps $f_A$ and $f_B$. We use the mutual correlation layer 
    to perform feature matching to obtain correspondence maps $C_{AB}$ and $C_{BA} $. 
    Then, using our Siamese decoder we predict the common objects masks $M_A$ and $M_B$.
    As shown before correlation layer, the heat-maps are covering all the objects inside the images. 
    After applying the correlation layer, the heat-maps on uncommon objects are  filtered out. Therefore, we
	utilize the output of the correlation layer to guide
	the network for segmenting the common objects.
    }
	\label{fig::corr_vis}
\end{figure}

The second part of our architecture is a mutual correlation layer. 
The outputs of encoders $f_A$ and $f_B$ represent the high-level semantic content of the input images. 
When the two images contain objects that belong to 
a common class, they should contain similar features at the locations 
of the shared objects. Therefore, we propose a mutual correlation 
layer to compute the correlation between each pair of locations on the 
feature maps. The idea of utilizing the correlation layer is inspired 
by Flownet \cite{Dosovitskiy_2015_ICCV}, in which the correlation layer 
is used to match feature points between frames for optical flow estimation.
Our motivation of using the correlation layer is to filter the
heat-maps (high-level features), which are generated separately for each input
image, to highlight the heat-maps on the common objects (see Fig.~\ref{fig::corr_vis}).
%
In detail, the mutual correlation layer performs a pixel-wise comparison 
between two feature maps $f_A$ and $f_B$. Given a point $(i,j)$ and 
a point $(m,n)$ inside a patch around $(i,j)$, the correlation between 
feature vectors $f_A(i,j)$ and $f_B(m,n)$ is defined as
\begin{equation}
C_{AB}(i,j,k) =\langle\ f_A(i,j),f_B(m,n)\rangle
\end{equation} 
where $k=(n-j)D+(m-i)$ and $D\times D$ is patch size. Since 
the common objects can locate at any place on the two input images, 
we set the patch size to $D=2*max(w-1,h-1)+1$, where $w$ and $h$ are 
the width and height of the feature maps $f_A$ and $f_B$.  
The output of the correlation layer is a feature map $C_{AB}$ of 
size  $w \times \ h \times D^2$. We use the same method to compute 
the correlation map $C_{BA}$ between $f_B$ and $f_A$.

\subsection{Siamese Decoder}
The Siamese decoder is the third part of our architecture, which predicts 
two foreground masks of the common objects. We squeeze the feature maps $f_A$ and $f_B$ and concatenate them with their correspondence maps $C_{AB}$ and $C_{BA}$ as the input to the Siamese decoder (Fig. \ref{fig:cnn_docs}).
The same as the Siamese encoder, the decoder is also 
arranged in a Siamese structure with shared parameters.
There are five blocks in our decoder, whereby each block has one 
deconvolutional layer and two convolutional layers. All the convolutional 
and deconvolutional layers in our Siamese decoder are followed by 
a ReLU activation function. By applying a Softmax function, the decoder 
produces two probability maps $p_A$ and $p_B$. Each probability map 
has two channels, background and foreground, with the same size 
as the input images.

\subsection{Loss Function}
\label{sub_sec::loss}
We define our object co-segmentation as a binary image labeling problem 
and use the standard cross entropy loss function to train our network. 
The full loss score $\mathcal{L}_{AB}$ is then estimated by 
$\mathcal{L}_{AB} =  \mathcal{L}_A + \mathcal{L}_B,$
where the $\mathcal{L}_A$ and the $\mathcal{L}_B$ are cross-entropy loss 
functions for the image $A$ and the image $B$, respectively. 

\subsection{Group Co-Segmentation}
\label{sub_sec::coseg_group}
Although our architecture is trained for image pairs, our method can handle a group of images. 
Given a set of 
$N$ images $\mathcal{I}=\left\lbrace I_1, ..., I_N \right\rbrace$,
we pair each image with \mbox{$K\leq N-1$} other images from $\mathcal{I}$.
Then, we use our DOCS network to predict the probability maps for the 
pairs, 
$\mathcal{P}=\left\lbrace p_n^k : 1\leq n \leq N , 1\leq k\leq K  \right\rbrace$, 
where $p_n^k$ is the predicted probability map for the $k$th pair of image $I_n$. 
Finally, we compute the final mask $M_n$ for image $I_n$ as
\begin{equation}
	M_n(x,y) = \mathrm{median} \{p_n^k(x,y) \} > \sigma.
\end{equation}
where $\sigma$ is the acceptance threshold. In this work, we set $\sigma=0.5$. We use the median to make our approach more robust to groups with outliers.

\section{Experiments}
\label{sec:experiments}

\subsection{Datasets}
\label{sec:datasets}
Training a CNN requires a lot of data. However, 
existing co-segmentation datasets are either too small or 
have a limited number of object classes. 
The MSRC dataset \cite{shotton2006textonboost} was first introduced 
for supervised semantic segmentation, then a subset was used for 
object co-segmentation \cite{vicente2011object}. This subset of MSRC 
only has 7 groups of images and each group has 10 images. 
The iCoseg dataset, introduced in \cite{batra2010icoseg}, consists 
of several groups of images and is widely used to evaluate 
co-segmentation methods. However, each group contains images of 
the same object instance or very similar objects from the same class.
The Internet dataset \cite{Rubinstein_2013_ICCV} contains thousands of 
images obtained from the Internet using image retrieval techniques. 
However, it only has three object classes: \textit{car, horse} and \textit{airplane}, 
where images of each class are mixed with other noise objects.
In \cite{Faktor_2013_ICCV}, Faktor and Irani use PASCAL 
dataset for object co-segmentation.  
They separate the images into 
$20$ groups according to the object classes and assume that each group 
only has one object. However, this assumption is not common for natural images.  

Inspired by \cite{Faktor_2013_ICCV}, we create an object co-segmentation 
dataset by adapting the PASCAL dataset labeled by \cite{BharathICCV2011}. 
The original dataset consists of $20$ foreground object classes and 
one background class. It contains $8,498$ training and $2,857$ validation 
pixel-level labeled images. From the 
training images, we sampled $161,229$ pairs of images, which 
have common objects, as a new co-segmentation training set. 
We used PASCAL validation images to sample $42,831$ validation pairs and $40,303$ test pairs.
Since our goal is to segment the common objects from the pair of images, 
we discard the object class labels and instead we label the common 
objects as foreground. Fig.~\ref{fig:example}(d) shows some examples of 
image pairs of our object co-segmentation dataset. In contrast to \cite{Faktor_2013_ICCV}, 
our dataset consists of image pairs of one or more arbitrary common classes.

\subsection{Implementation Details and Runtime}
\label{sec:training}
We use the Caffe framework \cite{jia2014caffe} 
to design and train our network. 
We use our co-segmentation dataset for training. We did not use 
any images from the MSCR, Internet or iCoseg datasets to fine tune our model.
The \textit{conv1-conv5} layers of our Siamese encoder (VGG-16 net \cite{VGG}) 
are initialized with weights trained on the Imagenet dataset \cite{imagenet}. 
We train our network on one GPU for $100$K iterations using Adam solver \cite{kingma2014adam}.
We use small mini-batches of $10$ image pairs, a momentum of $0.9$,  
a learning rate of $1e-5$, and a weight decay of $0.0005$.

Our method can handle a large set of images in linear time complexity $\mathcal{O}(N)$. As mentioned in Sec.~\ref{sub_sec::coseg_group} in order to co-segment an image, we pair it with $K$ ($K\leq N-1$) other images. In our experiments, we used all possible pairs to make the evaluations comparable to other approaches. Although in this case our time complexity is quadratic $\mathcal{O}(N^2)$, our method is significantly faster than others.

\vspace{-5mm}
\begin{table}[h]
\begin{center}
\begin{tabular}{c|r|r}
Number of images & Others time & Our time \\
\hline
$2$ & $8$ minutes \cite{joulin2010discriminative} & $0.1$ seconds\\
$30$ & $4$ to $9$ hours \cite{joulin2010discriminative} & $43.5$ seconds\\
$30$ & $22.5$ minutes \cite{Wang_2013_ICCV} & $43.5$ seconds\\
$418$ ($14$ categories, $\sim30$ images per category) & $29.2$ hours \cite{Faktor_2013_ICCV} & $10.15$ minutes\\
$418$ ($14$ categories, $\sim30$ images per category) & $8.5$ hours \cite{jerripothula2016image} & $10.15$ minutes\\
\end{tabular}
\end{center}
\vspace{-9mm}
\end{table}

\noindent
To show the influence of number of pairs $K$, we validate our method on the Internet dataset w.r.t. $K$ (Table~\ref{tab:k_influence}). Each image is paired with $K$ random images from the set.
As shown, we achieve state-of-the-art performance even with $K=10$. Therefore, the complexity of our approach is $\mathcal{O}(KN)=\mathcal{O}(N)$ which is linear with respect to the group size.

\setlength{\tabcolsep}{6pt}
\begin{table}[h]
\caption{Influence of number of pairs $K$.}
\label{tab:k_influence}
\begin{center}
\begin{tabular}{c|cc|cc|cc}
Internet & \multicolumn{2}{c|}{K=10} & \multicolumn{2}{c|}{K=20} & \multicolumn{2}{c}{K=99(all)}  \\
(N=100) & P & J & P & J & P & J\\
\hline
Car  & 93.93 & 82.89 & 93.91 & 82.85 & 93.90 & 82.81 \\ 
\hline
Horse & 92.31 & 69.12 & 92.35 & 69.17 & 92.45 & 69.44 \\
\hline
Airplane & 94.10 & 65.37 & 94.12 & 65.45 & 94.11 & 65.43\\
\hline
\textit{Average} & 93.45 & 72.46 & 93.46 & 72.49 & 93.49 & 72.56\\
\end{tabular}
\end{center}
\end{table}

\subsection{Results}
\label{sec:results}
We report the performance of our approach on MSCR 
\cite{shotton2006textonboost,vicente2010cosegmentation},
Internet \cite{Rubinstein_2013_ICCV}, and iCoseg \cite{batra2010icoseg} 
datasets, as well as our own co-segmentation dataset.

\subsubsection{Metrics.}
For evaluating the co-segmentation performance, there are two 
common metrics. The first one is \textit{Precision},
which is the percentage of correctly segmented pixels of both 
foreground and background masks. The second one is \textit{Jaccard},
which is the intersection over union of the co-segmentation result 
and the ground truth foreground segmentation. 

\begin{figure*}[t]
\begin{center}
\includegraphics[width=0.9\linewidth]{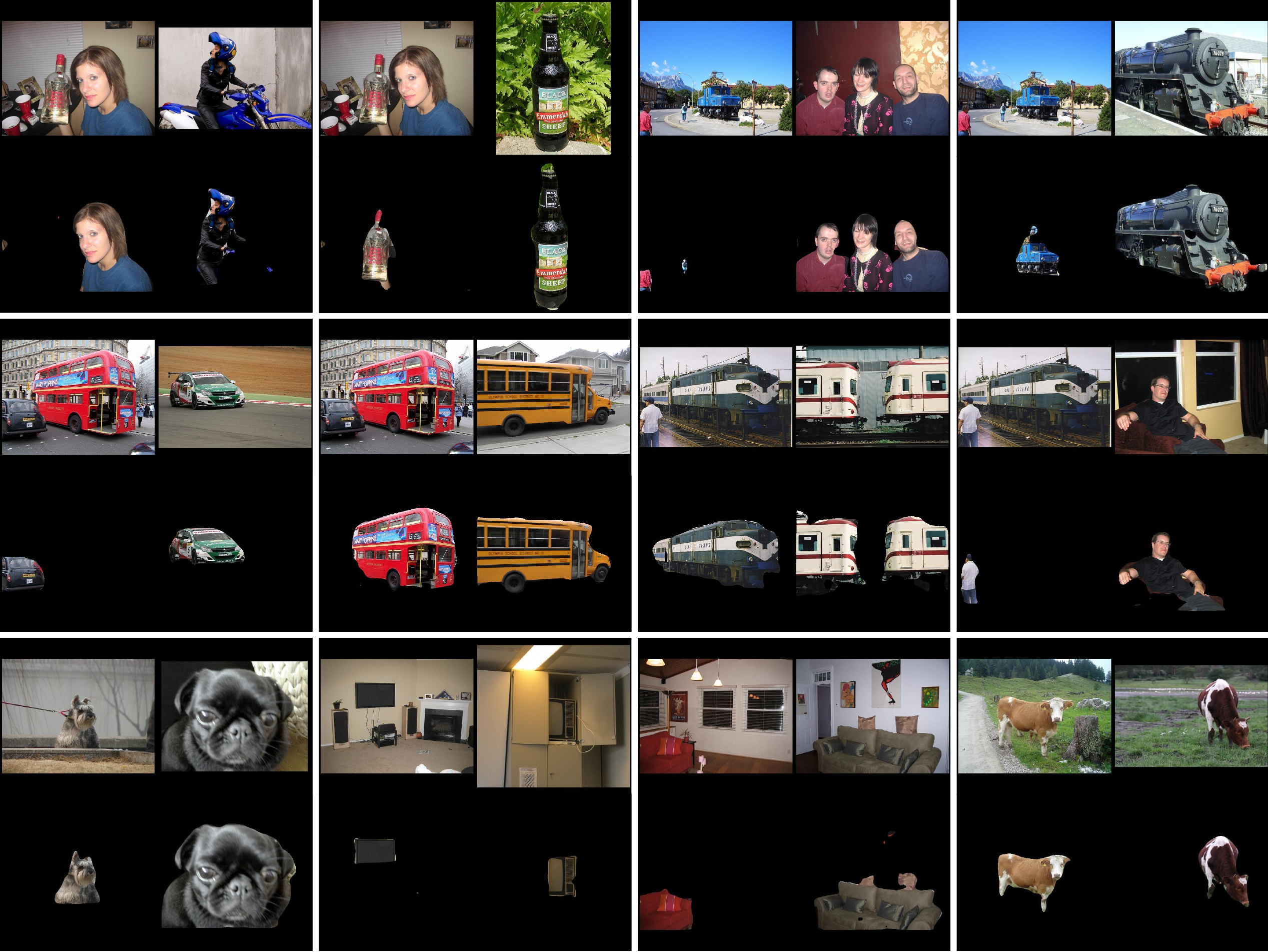}
\end{center}
   \caption{\textbf{Our qualitative results on PASCAL Co-segmentation dataset.}
   (odd rows) the input images, (even rows) the corresponding object 
   co-segmentation results.}
\label{fig:docs}
\end{figure*}

\subsubsection{PASCAL Co-Segmentation.}
As we mentioned in Sec~\ref{sec:datasets}, our co-segmentation dataset consists 
of $40,303$ test image pairs. 
We evaluate the performance of our method on our co-segmentation test data.
We also tried to obtain the common objects of same classes 
using a deep semantic segmentation model, here FCN8s \cite{Long_2015_CVPR}. 
First, we train FCN8s with the PASCAL dataset. 
Then, we obtain the common objects from two images by predicting 
the semantic labels using FCN8s and keeping the segments with 
common classes as foreground. Our co-segmentation method 
(\textbf{94.2\%} for \textit{Precision} and \textbf{64.5\%} for \textit{Jaccard}) 
outperforms FCN8s (\textbf{93.2\%} for \textit{Precision} and \textbf{55.2\%} 
for \textit{Jaccard}), which uses the same VGG encoder, and trained 
with the same training images. The improvement is probably 
due to the fact that our DOCS architecture is specifically designed 
for the object co-segmentation task, which FCN8s is designed for 
the semantic labeling problem. Another potential reason is that 
generating image pairs is a form of data augmentation. 
We would like to exploit these ideas in the future work. 
Fig. \ref{fig:docs} shows the qualitative results of our approach 
on the PASCAL co-segmentation dataset. We can see that our method successfully extracts 
different foreground objects for the left image when paired 
with a different image to the right.

\begin{figure}
\begin{center}
\includegraphics[width=0.90\linewidth, trim = {0 0 11.8cm 0}, clip]{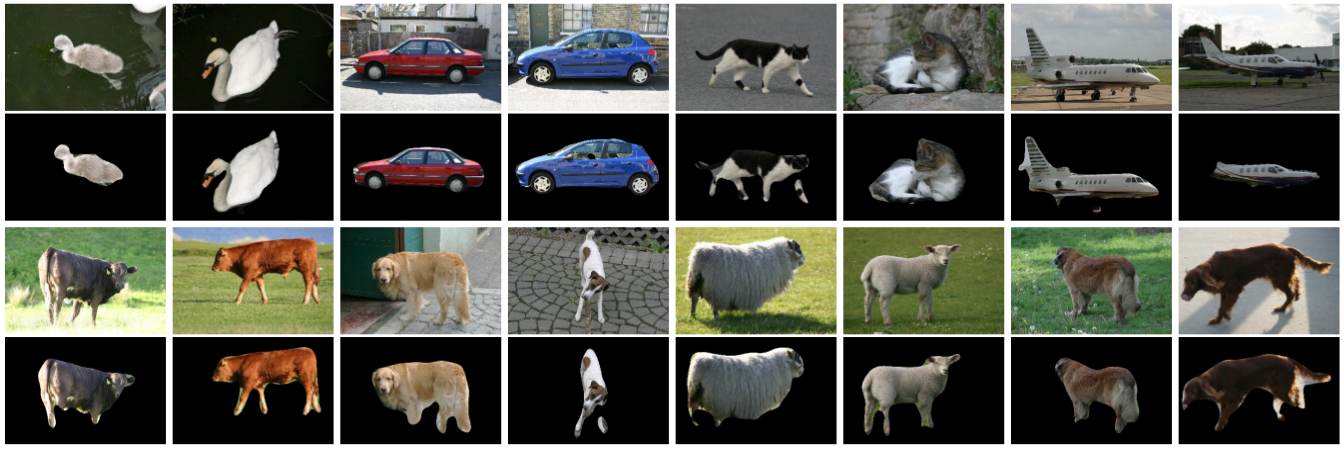}
\end{center}
   \caption{\textbf{Our qualitative results on the MSRC dataset (seen classes).} 
    (odd rows) the input images, (even rows) the corresponding object 
   co-segmentation results.}
\label{fig:MSRC}
\end{figure}

\setlength{\tabcolsep}{6pt}
\begin{table}[t]
\begin{center}
\caption{\textbf{Quantitative results on the MSRC dataset (seen classes).} 
	Quantitative comparison results of our DOCS approach with four 
    state-of-the-art co-segmentation methods on the co-segmentation 
    subset of the MSCR dataset.}
\label{tab:MSRC}
\begin{tabular}{c|cccccc}
MSCR  & \cite{vicente2011object} & \cite{Rubinstein_2013_ICCV} & \cite{Wang_2013_ICCV} & \cite{Faktor_2013_ICCV} & Ours\\[2pt]
\hline
Precision & 90.2 & 92.2 & 92.2  & 92.0 & \textbf{95.4} \Tstrut\\
Jaccard   & 70.6 & 74.7 &   -   & 77.0 & \textbf{82.9} \\
\end{tabular}
\end{center}
\end{table}
\setlength{\tabcolsep}{1.4pt}

\subsubsection{MSRC.}
The MSRC subset has been used to evaluate 
the object co-segmentation performance by many previous methods 
\cite{vicente2010cosegmentation,Rubinstein_2013_ICCV,Faktor_2013_ICCV,Wang_2013_ICCV}. For the fair comparison, we use the same subset as \cite{vicente2010cosegmentation}.
We use our group co-segmentation method to extract the foreground 
masks for each group. 
In Table. \ref{tab:MSRC}, we show the quantitative results of our method 
as well as four state-of-the-art methods 
\cite{vicente2011object,Rubinstein_2013_ICCV,Faktor_2013_ICCV,Wang_2013_ICCV}. 
Our \textit{Precision} and \textit{Jaccard} show a significant improvement 
compared to previous methods. 
It is important to note that \cite{vicente2011object} and 
\cite{Wang_2013_ICCV} are supervised methods, i.e.\ both use 
images of the MSRC dataset to train their models.
We obtain the new state-of-the-art results on this dataset 
even without training or fine-tuning on any images  from the 
MSRC dataset. Visual examples of object co-segmentation results 
on the subset of the MSRC dataset can be found in Fig.~\ref{fig:MSRC}. 

\begin{figure}
	\centering
    \includegraphics[width=0.90\linewidth]{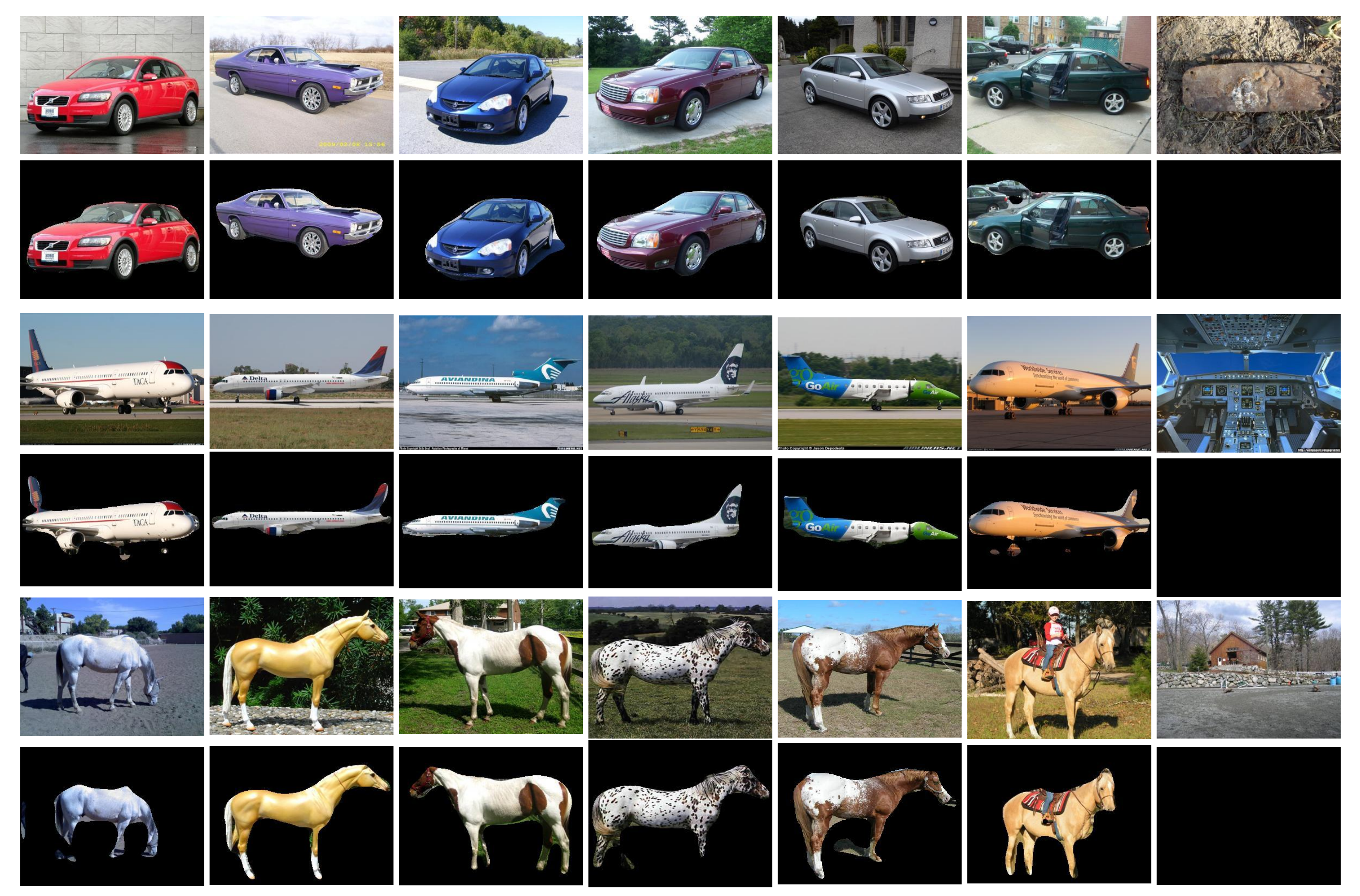}
   	\caption{\textbf{Our qualitative results on the Internet dataset (seen classes)}.  
   (odd rows) the input images, (even rows) the corresponding object 
   co-segmentation results.}
	\label{fig:internet}
\end{figure}

\setlength{\tabcolsep}{6pt}
\begin{table}
\begin{center}
\caption{\textbf{Quantitative results on the Internet dataset (seen classes).} 
Quantitative comparison of our DOCS approach with several state-of-the-art 
co-segmentation methods on the co-segmentation subset of the Internet dataset. 
`P' is the \textit{Precision}, and `J' is the \textit{Jaccard}.
}
\label{tab:Internet}
\begin{tabular}{c|c|cccccccc}
\multicolumn{2}{c|}{Internet} & \cite{joulin2010discriminative} & \cite{Rubinstein_2013_ICCV} & \cite{Chen_2014_CVPR} & \cite{Quan_2016_CVPR} & \cite{DDCRF} & Ours \\[2pt]
\hline
\multirow{2}{*}{Car}  & P  & 58.7  & 85.3 & 87.6 & 88.5 & 90.4 & \textbf{93.9} \Tstrut\\ 
 					  & J  & 37.1 & 64.4 & 64.9 & 66.8 & 72.0  & \textbf{82.8} \\[2pt] 
\hline
\multirow{2}{*}{Horse}  & P  & 63.8 & 82.8 & 86.2 & 89.3 & 90.2 & \textbf{92.4}  \Tstrut\\
 						& J  & 30.1 & 51.6 & 33.4 & 58.1 & 65.0 & \textbf{69.4}  \\[2pt]
\hline
\multirow{2}{*}{Airplane} & P  & 49.2 & 88.0 & 90.3 & 92.6 & 91.0 & \textbf{94.1} \Tstrut\\
 						  & J  & 15.3 & 55.8 & 40.3 & 56.3 & \textbf{66.0} & 65.4 \\[2pt]
\hline
\multirow{2}{*}{\textit{Average}} & P  & 57.2 & 85.4 & 88.0 & 89.6 & 91.1 & \textbf{93.5}\Tstrut\\
	 							  & J  & 27.5 & 57.3 & 46.2 & 60.4 & 67.7 & \textbf{72.6} \\
\end{tabular}
\end{center}
\end{table}
\setlength{\tabcolsep}{1.4pt}

\subsubsection{Internet.}
In our experiment, for the fair comparison, we followed 
\cite{Rubinstein_2013_ICCV,Chen_2014_CVPR,Quan_2016_CVPR,DDCRF}
to use the subset of the Internet dataset to evaluate our method. 
In this subset, there are 100 images in each category. 
We compare our method with five previous approaches 
\cite{joulin2010discriminative,Chen_2014_CVPR,Rubinstein_2013_ICCV,Quan_2016_CVPR,DDCRF}. 
Table \ref{tab:Internet} shows 
the quantitative results of each object category with respect to \textit{Precision}
and \textit{Jaccard}. We outperform most of the previous methods
\cite{joulin2010discriminative,Chen_2014_CVPR,Rubinstein_2013_ICCV,Quan_2016_CVPR,DDCRF} 
in terms of \textit{Precision} and \textit{Jaccard}. 
Note that \cite{DDCRF} is a supervised co-segmentation method, \cite{Chen_2014_CVPR}
trained a discriminative Latent-SVM detector and \cite{Quan_2016_CVPR} used a CNN trained on the ImageNet
to extract semantic features.
Fig. \ref{fig:internet} shows some quantitative results of our method. 
It can be seen that even for the `noise' 
images in each group, our method can successfully recognize them. 
We show the `noise' images in the last column.

\begin{figure*}[!htbp]
\begin{center}
\includegraphics[width=0.90\linewidth]{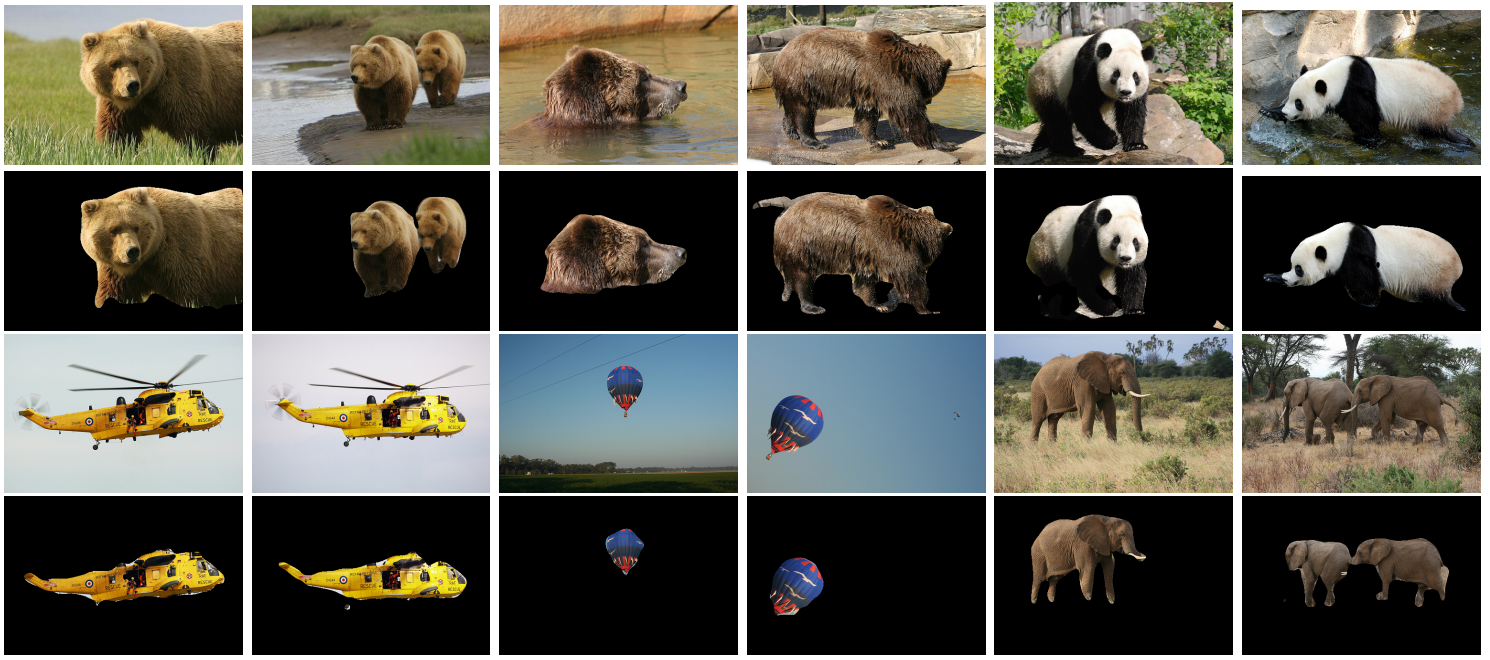}
\end{center}
   \caption{\textbf{Our qualitative results on iCoseg dataset (unseen classes).} 
   Some results of our object co-segmentation method, with input image pairs 
   in the odd rows and the corresponding object co-segmentation results in 
   the even rows. For this dataset, the object classes were not known 
   during training of our method (i.e. \textit{unseen}).}
\label{fig:icoseg}
\end{figure*}

\setlength{\tabcolsep}{6pt}
\begin{table}[!h]
  \begin{center}
  \caption{\textbf{Quantitative results on the iCoseg dataset (unseen classes).} 
  Quantitative comparison of our DOCS approach with four state-of-the-art 
  co-segmentation methods on some object classes of the iCoseg dataset, 
  in terms of Jaccard. For this dataset, these object classes were not known 
  during training of our method (i.e. \textit{unseen}).}
  \label{tab:iCoseg}
  \begin{tabular}{l|ccccccccc}
  iCoseg & \cite{Rubinstein_2013_ICCV} & \cite{jerripothula2014automatic} &\cite{Faktor_2013_ICCV} &\cite{jerripothula2016image}  & Ours \\[2pt]
  \hline
  bear2		& 65.3 	& 70.1	& 72.0 & 67.5 & \textbf{88.7} \Tstrut\\ 
  brownbear	& 73.6 	& 66.2 	& \textbf{92.0} & 72.5 & 91.5 \\
  cheetah 	& 69.7 	& 75.4 & 67.0 & \textbf{78.0}  & 71.5 \\ 
  elephant 	& 68.8 	& 73.5 	& 67.0 	& 79.9 	& \textbf{85.1} \\ 
  helicopter 	& 80.3 	& 76.6 	& \textbf{82.0} &80.0  & 73.1 \\
  hotballoon  & 65.7 	& 76.3  & 88.0 & 80.2 & \textbf{91.1} \\ 
  panda1 		& 75.9 	& 80.6 & 70.0 & 72.2 & \textbf{87.5} \\ 
  panda2 		& 62.5 	& 71.8 & 55.0 & 61.4 & \textbf{84.7} \\[2pt]  
  \hline
  \textit{average} & 70.2  & 73.8 & 78.2 & 74.0 & \textbf{84.2} \Tstrut\\ 
  \end{tabular}
  \end{center}
\end{table}
\setlength{\tabcolsep}{1.4pt}

\subsubsection{iCoseg}
To show that our method can generalize on \textit{unseen classes}, 
\ie classes which are not part of the training data, we need to 
evaluate our method on \textit{unseen classes}.
Batra \etal \cite{batra2010icoseg} introduced the iCoseg dataset
for the \textit{interactive} co-segmentation task. 
In contrast to the MSRC and Internet datasets, 
there are multiple object classes in the iCoseg dataset 
which do not appear in PASCAL VOC dataset. Therefore, it is possible to use 
the iCoseg dataset to evaluate the generalization of 
our method on {\it unseen object classes}. 
We choose eight groups of images from the iCoseg dataset 
as our unseen object classes, 
which are \textit{bear2, brown\_bear, cheetah, elephant, helicopter, hotballoon, panda1} and \textit{panda2}. 
There are two reasons for this choice: firstly, these object classes 
are not included  in the PASCAL VOC dataset. 
Secondly, in order to focus on  \textit{objects}, in contrast to \textit{stuff}, 
we ignore groups like \textit{pyramid, stonehenge} and \textit{taj-mahal}. 
We compare our method with four state-of-the-art approaches 
\cite{jerripothula2014automatic,Rubinstein_2013_ICCV,Faktor_2013_ICCV,jerripothula2016image} 
on unseen objects of the iCoseg dataset. 
Table~\ref{tab:iCoseg} shows the comparison results of each unseen 
object groups in terms of \textit{Jaccard}. The results show that 
for 5 out of 8 object groups our method performs best, and 
it is also superior on average.  Note that the results of 
\cite{jerripothula2014automatic,Rubinstein_2013_ICCV,Faktor_2013_ICCV,jerripothula2016image} 
are taken from Table X in \cite{jerripothula2016image}.
Fig. \ref{fig:icoseg} shows some qualitative results of our method. 
It can be seen that our object co-segmentation method can detect 
and segment the common objects of these unseen classes accurately.

Furthermore to show the effect of number of PASCAL classes on the performance of our approach on unseen classes, we train our network on partial randomly picked PASCAL classes, \ie $\lbrace5, 10, 15\rbrace$, and evaluate it on the iCoseg unseen classes. As it is shown in Table~\ref{tab:pascal_class}, our approach can generalize to unseen classes even when it is trained with only 10 classes from PASCAL.

\setlength{\tabcolsep}{6pt}
\begin{table}[h]
  \caption{Analyzing the effect of number of training classes on unseen classes.}
  \label{tab:pascal_class}
  \centering
  \begin{tabular}{l|ccccc}
  iCoseg & P(5) & P(10) & P(15) & P(20) \\[2pt]
  \hline
  \textit{average} & 75.5 & 83.9 & 83.7 & 84.2 \\
  \end{tabular}
\end{table}

\subsection{Ablation Study}
To show the impact of the mutual correlation layer in our network architecture,
we design a baseline network \textit{DOCS-Concat} without using mutual correlation layers.
In detail, we removed the correlation layer and we concatenate $f_A$ and $f_B$ (instead of $C_{AB}$) for image $I_A$ and concatenate $f_B$ and $f_A$ (instead of $C_{BA}$) for image $I_B$. 
In Table~\ref{tab:ablation}, we compare the performance of different network designs on multiple datasets. As shown, the mutual correlation layer in \textit{DOCS-Corr} improved the performance significantly.

\begin{table}
  \caption{\textbf{Impact of mutual correlation layer.}}
  \label{tab:ablation}
  \centering
  \begin{tabular}{c|cc|cc}
       & \multicolumn{2}{c|}{DOCS-Concat} & \multicolumn{2}{c}{DOCS-Corr} \\
       & Precision & Jaccard & Precision & Jaccard \\[2pt]
       \hline 
       Pascal VOC 	  & 92.6 & 49.9 & \textbf{94.2} & \textbf{64.5}  \Tstrut\\
       MSRC 	  	  & 92.6 & 72.0 & \textbf{95.4} & \textbf{82.9}  \\
       Internet   	  & 91.8 & 62.7 & \textbf{93.5} & \textbf{72.6}  \\
       iCoseg(unseen) & 93.6 & 78.9 & \textbf{95.1} & \textbf{84.2}  \\
  \end{tabular}
\end{table}

\section{Conclusions}
In this work, we presented a new and efficient CNN-based method for 
solving the problem of object class co-segmentation, which consists of 
jointly detecting and segmenting objects belonging to a common semantic 
class from a pair of images. Based on a simple encoder-decoder architecture, 
combined with the mutual correlation layer for matching semantic features, 
we achieve state-of-the-art performance on various datasets, and demonstrate 
good generalization performance on segmenting objects of new semantic classes, 
unseen during training. To train our model, we compile a large object 
co-segmentation dataset consisting of image pairs from PASCAL dataset 
with shared objects masks.

\paragraph{}
\textbf{Acknowledgements} This work is funded by the DFG grant “COVMAP: Intelligente Karten mittels gemeinsamer GPS- und Videodatenanalyse” (RO 4804/2-1).

%
%
%
\bibliographystyle{splncs04}
\bibliography{docsbib}
\end{document}